\documentclass[letterpaper, 10 pt, conference]{ieeeconf}  % Comment this line out if you need a4paper

\IEEEoverridecommandlockouts                              % This command is only needed if
\usepackage{graphics} % for pdf, bitmapped graphics files
\usepackage{amsmath} % assumes amsmath package installed
\usepackage{amssymb}  % assumes amsmath package installed
\usepackage[colorlinks, linkcolor=blue, citecolor=blue]{hyperref}
\usepackage{cite}
\usepackage{amssymb}
\usepackage{graphicx}
\usepackage{caption}
\usepackage{subfigure}
\usepackage{mathrsfs}
\usepackage{array}
\usepackage{multirow}
\usepackage{makecell}
\usepackage{xcolor}
\usepackage{amsfonts}
\usepackage{gensymb}
\usepackage{threeparttable}
\makeatletter
\renewcommand*\env@matrix[1][*\c@MaxMatrixCols c]{%
 \hskip -\arraycolsep
 \let\@ifnextchar\new@ifnextchar
 \array{#1}}
\makeatother

\title{\LARGE \bf
Robust localization for planar moving robot in changing environment: \\A perspective on density of correspondence and depth
}
\author{Yanmei Jiao$^{1}$, Lilu Liu$^{1}$, Bo Fu$^{1}$, Xiaqing Ding$^{1}$, Minhang~Wang$^{2}$, Yue Wang$^{1}$ and Rong Xiong$^{1}$% <-this % stops a space
\thanks{$^{1}$Yanmei Jiao, Lilu Liu, Bo Fu, Xiaqing Ding, Yue Wang and Rong Xiong are with the State Key Laboratory of Industrial Control and Technology, Zhejiang University, Hangzhou, P.R. China. $^{2}$Minhang Wang is with the Application Lab, Huawei Incorporated Company, P.R. China. Yue Wang is the corresponding author {\tt\small wangyue@iipc.zju.edu.cn}.}%
}

\begin{document}

\maketitle
\thispagestyle{empty}
\pagestyle{empty}

%%%%%%%%%%%%%%%%%%%%%%%%%%%%%%%%%%%%%%%%%%%%%%%%%%%%%%%%%%%%%%%%%%%%%%%%%%%%%%%%

\begin{abstract}
Visual localization for planar moving robot is important to various indoor service robotic applications. To handle the textureless areas and frequent human activities in indoor environments, a novel robust visual localization algorithm which leverages dense correspondence and sparse depth for planar moving robot is proposed. The key component is a minimal solution which computes the absolute camera pose with one 3D-2D correspondence and one 2D-2D correspondence. The advantages are obvious in two aspects. First, the robustness is enhanced as the sample set for pose estimation is maximal by utilizing all correspondences with or without depth. Second, no extra effort for dense map construction is required to exploit dense correspondences for handling textureless and repetitive texture scenes. That is meaningful as building a dense map is computational expensive especially in large scale. Moreover, a probabilistic analysis among different solutions is presented and an automatic solution selection mechanism is designed to maximize the success rate by selecting appropriate solutions in different environmental characteristics. Finally, a complete visual localization pipeline considering situations from the perspective of correspondence and depth density is summarized and validated on both simulation and public real-world indoor localization dataset. The code is released on github\footnote{\url{https://github.com/slinkle/1P1DP}}.
\end{abstract}

%the sample set for pose estimation is maximal as all matched correspondences with or without depth can be utilized, which directly promotes the robustness. Second, dense matches rather than sparse matches can be exploited to handle textureless and repetitive texture scenes without extra effort to build the dense map.

%%%%%%%%%%%%%%%%%%%%%%%%%%%%%%%%%%%%%%%%%%%%%%%%%%%%%%%%%%%%%%%%%%%%%%%%%%%%%%%%
\section{Introduction}

Indoor service robots have been applied in various scenarios during the last decade, such as home, office, restaurant and so on\cite{shi2020we}\cite{ruiz2017robot}. One of the fundamental techniques for this success is the 2D LiDAR based localization of planar moving robots \cite{thrun2002probabilistic}. To further reduce the cost, cameras are expected to provide visual localization for service robots. However, due to the sensitivity to frequent and complex environmental changes e.g. illumination, texture, objects presence, reliable visual indoor localization remains a challenge \cite{taira2018inloc}.

%4-points 3-points [mono and multi] 2-points [kneip kukelova jiao] 1point1line[jiao]
%讲讲室内的motion特点，可以利用，讲一些利用motion的ransac工作，和鲁棒性，对那些环境和视角的change有好处。这里所有这些工作主要在scsd上
%此外，室内要更多feature 引inloc  a robust estimator such as RANSAC \cite{fischler1981random}\cite{choi1997performance}
The common pipeline for visual localization is to establish the feature correspondences from the query image to the map and recover the 6 degree of freedom (DoF) camera pose through geometric estimation. In this pipeline, feature matching is vulnerable to environmental changes, causing presence of outliers. To relieve this problem, minimal solutions exploiting minimal number of correspondences for pose estimation call for research, which can be embedded into RANSAC \cite{fischler1981random}\cite{choi1997performance} to improve the robustness. For 6DoF solution, 3 correspondences are minimally required for both mono-camera \cite{gao2003complete}\cite{lepetit2009epnp} and multi-camera scenarios \cite{kneip2013using}\cite{kneip2014upnp}. Moreover, when the inertial measurement is available, the direction of gravity is known, further reducing the minimal number of correspondences to 2 as shown in \cite{kukelova2010closed}\cite{jiao20202}. Note that solutions above are mainly applied in the situations where sparse correspondences and sparse depth (SCSD) are available. Whereas in indoor localization, only sparse features \cite{lowe2004distinctive}\cite{rublee2011orb}\cite{detone2018superpoint} may not perform well due to the textureless and highly repetitive areas (\textit{e.g.}, walls, windows and floors) \cite{taira2018inloc}.
%  and the short viewing distance will cause larger appearance changes when varying viewpoints

\begin{figure}[tbp]
    \centering
    \includegraphics[width=0.5\textwidth]{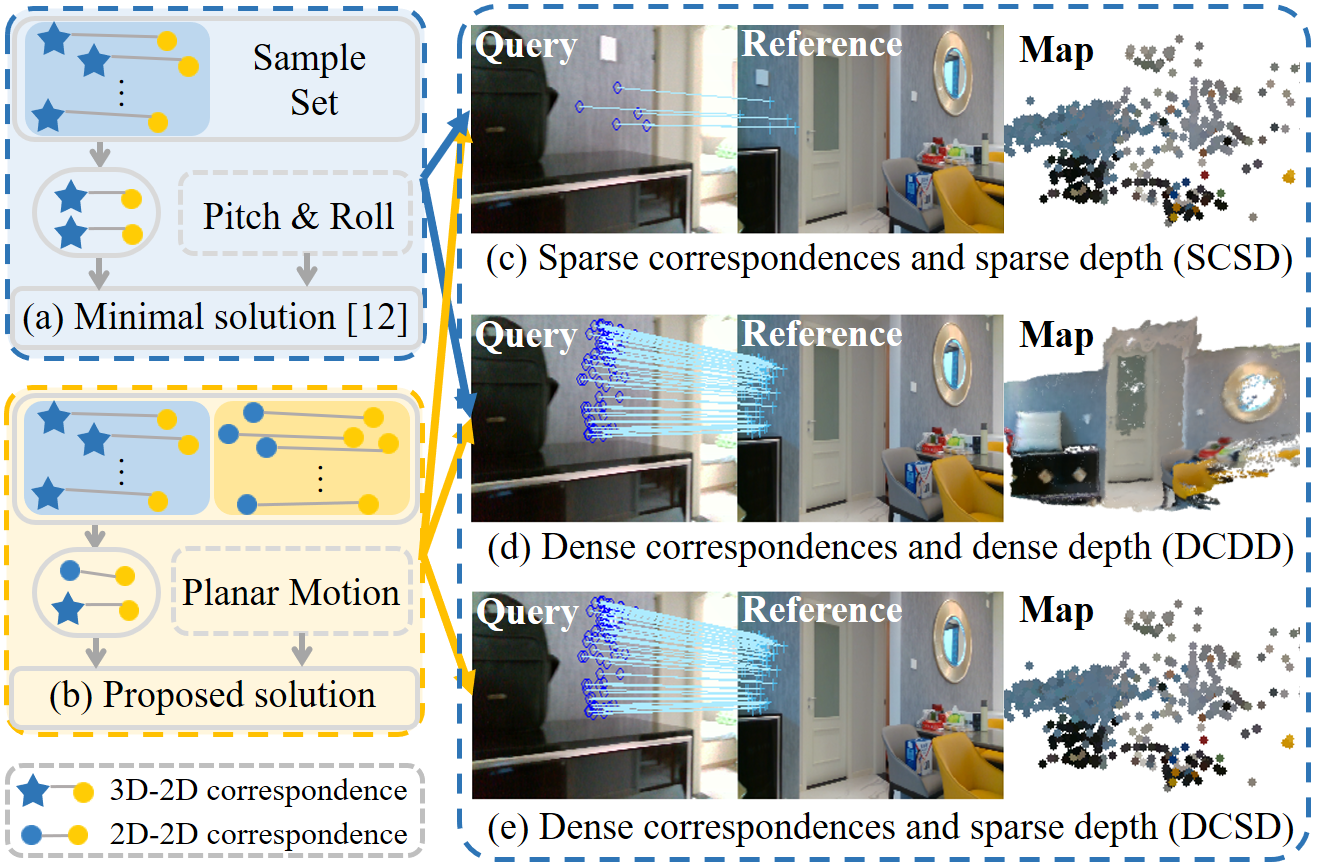}
    \caption{Left: The sample set comparison of (a) 3D-2D feature based minimal solution and (b) the proposed minimal solution. Right: The illustration of different settings (c) SCSD, (d) DCDD and (e) DCSD. The proposed solution is the first to deal with DCSD as all correspondences with or without depth can be utilized.}
    \label{fig.overview}
\end{figure}
%The proposed solution is the first to deal with DCSD such that 
%All matched features (with or without depth) can be used for pose estimation with the proposed solution, which also enables to utilize dense matches while maintaining a sparse map.
% dnn带来更多的correspondence可能, dc如果用传统方法，会引出dcdd的mapping问题；另一个视角是dcsd问题。
% 引出研究问题 dcsd，以及基于此对dcdd和scsd的场景进行探讨，方法的选择         把问题写的更清楚和highlight一点
In recent years, the development of deep neural network brings the possibility for efficiently matching all pixels to generate dense correspondences \cite{rocco2020ncnet}. In \cite{taira2018inloc}, with lots of correspondences, the conventional minimal solution based RANSAC is applied to eliminate the large percentage of outliers, achieving superior accuracy. We argue that this pipeline actually solves the problem of dense correspondences and dense depth (DCDD), which requires a dense 3D model in prior e.g. 3D scanning in \cite{taira2018inloc}. However, such model is difficult to build as it requires large computational capacity and can be noisy. A more applicable setting is localization with dense correspondences but sparse depth (DCSD), which only requires a conventional sparse environmental model built by sparse bundle adjustment \cite{mur2015orb}\cite{schneider2018maplab}, as shown in Fig. \ref{fig.overview}. To solve DCSD problem, still utilizing the conventional pipeline wastes those inliers among dense correspondences without depth, which should have made contributions to outliers elimination. An ideal pipeline for DCSD is to use both 2D-3D feature matches (correspondences with depth) and 2D-2D feature matches (correspondences without depth) in RANSAC to improve robustness and accuracy.

% 我们把motion prior思路用到dcsd里，提出一种新的最小解，以及对应的pose estimator, 再分析dcsd scsd dcdd，去指导三个situation
% 最后提出一个pipe
%In this paper, we propose a minimal solution utilizing only 2 correspondences, which can either be 2D-3D correspondences, or 2D-2D correspondence, by taking planar motion constraint into consideration. Leveraging this minimal solution, we further discuss the utilization of these minimal solutions in settings with SCSD, DCSD and DCDD, and propose a completed pipeline for solving robust visual localization, which we consider is valuable for service robots in changing environments.
%By composing these modules together, we propose a completed robust visual localization pipeline to achieve the best success rate.

In this paper, we propose a minimal solution utilizing one 2D-3D correspondence combined one 2D-2D correspondence by taking planar motion constraint into consideration. Specifically, according to the geometry, one point correspondence with depth (1DP) provides two independent constraints about the pose, and one point correspondence without depth (1P) provides one constraint, leading to the 1P1DP based minimal solution for the 3DoF planar motion. To the best of our knowledge, this is the first minimal solution designed for planar motion given DCSD. In addition, together with our previous work on minimal solution with two correspondences with depth (2DP) \cite{jiao20202}, we present a probabilistic analysis on the two solutions given different outlier ratio and reliable depth ratio, which guides the design of a solution selection mechanism. By composing these modules together, we further discuss the utilization of different minimal solutions in settings with SCSD, DCSD and DCDD, and propose a completed pipeline for robust visual localization. In summary, the contributions of this paper are presented as follows:
%, in which the number of feature correspondences is minimal (only two) and the utilization of structure information is smallest (only one depth)
\begin{itemize}
\item A minimal closed-form solution namely 1P1DP in mono-camera system and MC1P1DP in multi-camera system is proposed for absolute pose estimation of planar moving robot.
\item A probabilistic analysis on 1P1DP and 2DP given different environmental characteristics, SCSD, DCSD and DCDD, is presented to guide the design of a solution selection mechanism.
\item A completed pipeline for robust visual localization is proposed to automatically pick the appropriate solution according to the current environment, which is then validated on lifelong indoor localization datasets.
\end{itemize}

\section{Related Works}
% features  \noindent
\subsection{Density of feature correspondences}
Lots of efforts have been paid to design the robust feature detectors and descriptors in geometric computer vision community. The typical pipeline for handcrafted features is to detect a keypoint and then describe it \cite{lowe2004distinctive}\cite{leutenegger2011brisk}. Besides robustness, the real time property is also required for features \cite{viswanathan2009features}\cite{rublee2011orb}. As convolutional neural networks show superior performance on representation, many learning-based feature detectors \cite{savinov2017quad}\cite{zhang2018learning}\cite{barroso2019key} and descriptors \cite{balntas2016learning}\cite{simo2015discriminative}\cite{simonyan2014learning} are proposed to replace the handcrafted procedures. With the improvement of the computing capability, dense feature descriptors can be learned along with detectors to encode more information about the detected keypoints\cite{detone2018superpoint}\cite{revaud2019r2d2}. Then dense feature detectors which perform detection densely on whole image and generate matches from pixel to pixel also developed \cite{rocco2020ncnet}. This approach often capture global information for feature matching and has shown to provide better correspondences especially in indoor localization\cite{taira2018inloc}. However, the construction of dense 3D map is not mature as sparse map \cite{schonberger2016structure}\cite{mur2015orb}\cite{schneider2018maplab} as it requires huge computation and storage capacity, which significantly limits the application of dense features in localization. Therefore, considering the difficulty of the dense map construction and the advantage of dense features in indoor localization, the combination of dense correspondences and sparse depth is valuable.

% minimal solutions
\subsection{Robust pose estimation solutions}
Estimating the camera pose with the feature matches and 3D map as input is typically performed by PnP solvers \cite{gao2003complete}\cite{lepetit2009epnp}\cite{jiao20202} embedded in robust estimator such as RANSAC \cite{fischler1981random}\cite{choi1997performance}. However, these solutions are designed only for features with depth. Therefore, the existing 2D-3D pose estimation algorithms can only cooperate with SCSD or DCDD. The other option to use dense correspondences without dense depth is to compute the pose only with 2D-2D correspondences via epipolar geometry\cite{zhou2020learn}. There are also many minimal solvers for the pose estimation with pure 2D-2D correspondences, such as classical 8-point \cite{hartley2003multiple} and 5-point \cite{nister2004efficient} and \cite{fraundorfer2010minimal}\cite{chou20152} for reduced DoF situation. Although experiments show that it's practical for localization without depth information, it's inaccurate as it's difficult to constraint scale in the pose computation or optimization. In summary, the solutions for pose estimation are designed only for 2D-3D or 2D-2D correspondences. Therefore, a new solution is needed for utilizing both 2D-3D correspondences obtained from features matched with sparse map and 2D-2D correspondences obtained from the other dense features in DCSD, which is the focus of this paper.

\section{Minimal Solution}
\begin{figure}[tbp]
    \centering
    \includegraphics[width=0.4\textwidth]{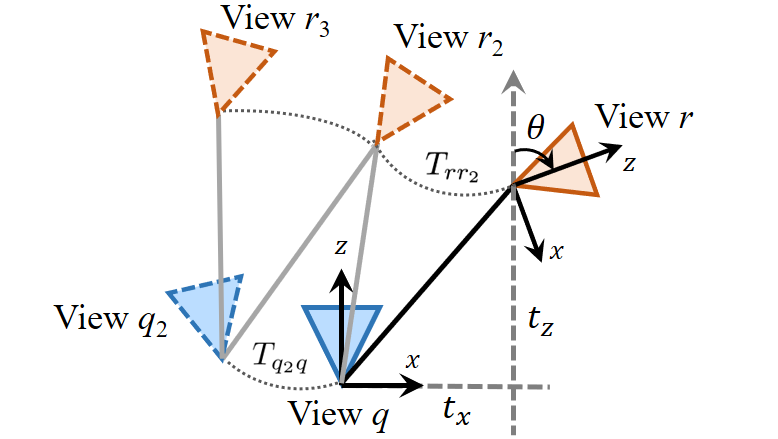}
    \caption{The illustration of planar motion between multiple query and reference views.}
    \label{fig.frames}
  \vspace{-0.3cm}
\end{figure}

In this paper, the planar motion property is employed to formulate the indoor localization problem. We assume the image plane is vertical to the ground such that the unknown variables of the pose between reference view $r$ and query view $q$ are the translation along axis $x$ and $z$ and the rotation around $y$, which are denoted as $t_x$, $t_z$ and $\theta$ as illustrated in Fig. \ref{fig.frames}. Note that the assumption is easy to satisfy by rotating the appropriate pitch and roll angles obtained from calibration parameters or inertial measurements. Then the rotation and translation matrix from camera coordinate system of view $r$ to view $q$ can be written as:
\begin{equation}\label{eq.Rt}
R=\begin{bmatrix}
                cos(\theta)& 0 & sin(\theta) \\
               0  & 1 & 0 \\
               -sin(\theta) & 0 & cos(\theta)
             \end{bmatrix}, t={\begin{bmatrix}
      t_x \\
      0 \\
      t_z
    \end{bmatrix}}
\end{equation}

%\noindent
\textbf{Constraints from 2D-3D correspondence}: Given a 2D feature expressed as $p_1=[\tilde{u}_{1},\tilde{v}_{1},1]^T=K^{-1}[{u}_{1},{v}_{1},1]^T$ in view $q$ and the corresponding 3D point $P_1=[x_1, y_1, z_1]^T$ in camera coordinate system of view $r$, according to the projection geometry, we have:
\begin{equation}\label{2d-3d}
  \frac{R_1 P_1 + t_x}{\tilde{u}_{1}} = \frac{R_2 P_1}{\tilde{v}_{1}} = R_3 P_1 + t_z
\end{equation}
where $R\triangleq [R_1^T,R_2^T,R_3^T]^T$ and $K$ is the camera intrinsic parameters. Two constraints can be derived from (\ref{2d-3d}) as:
\begin{eqnarray}\label{con1}
\tilde{v}_{1} x_1 cos(\theta) + \tilde{v}_{1} z_1 sin(\theta) + \tilde{v}_{1} t_x - \tilde{u}_{1} y_1 =0 \\ 
\tilde{v}_{1} z_1 cos(\theta) - \tilde{v}_{1} x_1 sin(\theta) + \tilde{v}_{1} t_z - y_1 =0 \label{con2}
\end{eqnarray}

%\begin{equation}\label{con1}
%\tilde{v}_{1} x_1 cos(\theta) + \tilde{v}_{1} z_1 sin(\theta) + \tilde{v}_{1} t_x - \tilde{u}_{1} y_1 =0
%\end{equation}
%\begin{equation}\label{con2}
%\tilde{v}_{1} z_1 cos(\theta) - \tilde{v}_{1} x_1 sin(\theta) + \tilde{v}_{1} t_z - y_1 =0
%\end{equation}

% (\ref{con1}) - (\ref{con2})
It's clear that given another 2D-3D correspondence, two constraints similar to (\ref{con1}) - (\ref{con2}) can be obtained. The pose can be solved by least square with the four constraints. Since the number of constraints provided by this solution is larger than the unknowns, it can achieve higher accuracy.
%As the planar localization problem is 3DoF, the number of constraints provided by this solution is larger than the unknowns, such that it's 

%\noindent
\textbf{Constraints from 2D-2D correspondence}: Denoting the 2D-2D correspondence as $p_2$ from view $q$ and $p_3$ from view $r$, the epipolar constraint can be expressed as:
\begin{equation}\label{con3}
  {p_2}^T [t]_{\times} R p_3 = 0
\end{equation}
where $[\ \cdot \ ]_{\times}$ denotes the skew symmetry matrix.

Representing unknowns $t_x$ and $t_z$ by $\theta$ from (\ref{con1}) - (\ref{con2}) and substituting them into (\ref{con3}), the pose can be solved and the solution is denoted as 1P1DP. The solution is the first to combine correspondence with and without depth for pose estimation in planar motion localization problem.
%, which is the minimal solution for planar motion visual localization problem

%\noindent
\textbf{Extending to multi-reference case}: As one query image may correspond to multiple reference images in map \cite{arandjelovic2016netvlad}, integrating these correspondences from other reference images will promote the localization performance. Suppose to sample the 2D-2D correspondence between view $r_2$ and view $q$, which is different from that provides the 2D-3D correspondence. As the camera poses of reference views can be obtained from the map, the relative transformation between the two reference views $T_{rr_2}$ is known. Then we have:
\begin{equation}\label{transform}
T_{qr_2}=T_{qr}T_{rr_2}=\begin{bmatrix}
                        R & t \\
                        \textbf{0} & 1
                      \end{bmatrix} T_{rr_2}
\end{equation}

Note that the unknowns in $T_{qr_2}$ are the same as in $T_{qr}$, so according to the epipolar constraint:
\begin{equation}\label{2d-2d}
  {p_2}^T [t_{qr_2}]_{\times} R_{qr_2} p_3 = 0
\end{equation}
the 1P1DP for multi-reference case can also be solved.

%\noindent
\textbf{Extending to multi-camera system}: The proposed solution can be easily extended to physical multi-camera system or temporal multi-query from continuous frames, such that more observations from other query views can assist the pose estimation of current query view. As shown in Fig. \ref{fig.frames}, the relative transformation between the two query views $T_{q_2q}$ can be obtained by extrinsic parameters in multi-camera system or visual odometry in multi-query case.

Considering one 2D-2D correspondence provided by view $q_2$ and $r_2$, the (\ref{transform}) becomes
\begin{equation}\label{transform2}
T_{q_2r_2}=T_{q_2q}T_{qr}T_{rr_2}
\end{equation}
And the constraint can be obtained by substituting $R_{q_2r_2}$ and $t_{q_2r_2}$ into (\ref{2d-2d}). The rest is similar to the previous 1P1DP solution. The solution for multi-camera system is denoted as MC1P1DP.

\section{Solution Selection}\label{model}

To deal with outliers, the proposed 1P1DP minimal solution is embedded into RANSAC \cite{fischler1981random} to achieve robust visual localization. In our previous work, we also propose another 2 points based minimal solution, 2DP \cite{jiao20202}. To pick the appropriate one in real application, we theoretically analyze the performance of 1P1DP and 2DP with respect to environmental characteristics for guidance.

\subsection{Success rate comparison}

Recalling the procedure of obtaining the 2D-3D correspondences, we first retrieve the 2D-2D correspondences between the query image and the reference image, then extract the corresponding 3D information of the reference image in map. In the following, we first consider SCSD, under which the number of 2D-2D and 2D-3D correspondences are the same.

\textbf{SCSD:} Denote the inlier rate of the 2D-2D correspondences as $\lambda$. As the depth measurement can be noisy due to distance and material, we also denote the reliable depth rate of reference features as $\gamma$, $(0 \leq \lambda , \gamma\leq 1)$. Then the success rate of one trial in RANSAC with 1P1DP solution is:
\begin{equation}\label{suc-1p1dp}
P_{SCSD-1P1DP}=\lambda\cdot(\lambda \gamma)
\end{equation}
and the success rate of one 2DP sample is:
\begin{equation}\label{suc-2dp}
P_{2DP}=(\lambda \gamma)\cdot(\lambda \gamma)
\end{equation}
We have $P_{2DP}\leq P_{SCSD-1P1DP}$ and the equality exists when $\gamma=1$, which indicates the difference of the success rate between two solutions depends on the reliable depth rate of the features in the reference image. If the reliable depth rate over the whole reference image is high, the success rate of 2DP is similar to that of 1P1DP. Note that the constraints provided by 2DP is more than that of 1P1DP and the accuracy of 2DP is also better than 1P1DP. However, when reliable depth rate is low, 1P1DP will show advantage, say most features are detected in a distant picture on the wall.

%To measure the depth reliability, we consider three metrics as shown in Fig. \ref{fig.example}: the distance of the point to the camera, the uncertainty of the depth of each point after sparse bundle adjustment and the total observation number of the point in the map.

\textbf{DCSD:} In this situation, the inlier rate of 2D-2D correspondences is different. Denote the inlier rate of dense 2D-2D correspondences as $\lambda_d$, we have
\begin{equation}\label{suc-Dense1p1dp}
P_{DCSD-1P1DP}=\lambda_d\cdot(\lambda \gamma)
\end{equation}
Comparing with (\ref{suc-Dense1p1dp}) and (\ref{suc-1p1dp}), the difference depends on the inlier rate of dense and sparse correspondences. When $\lambda_d>\lambda$, the advantage of 1P1DP in DCSD becomes more obvious. Note that dense correspondences are usually generated by networks that considers neighborhood consistency, higher $\lambda_d$ in DCSD is thus practical.

\textbf{DCDD:} In this situation, $\gamma$ significantly increases. Therefore, as in SCSD with high $\gamma$, 2DP can be the best choice given its better accuracy.

We find that 1P1DP and 2DP have advantages respectively given different environmental characteristics, inspiring that a better performance can be achieved if the solution is picked correctly for complement. In summary, based on the theoretic analysis as guidance, we present a robust visual localization pipeline from the perspective on density of correspondence and depth as shown in Fig. \ref{fig.framework}.

%After above analysis, an automatic solution selection strategy is needed to combine the advantage of 2DP and Dense1P1DP according to different environmental characteristics, which can achieve the best accuracy and robustness under the sparse depth map.
%In the next subsection, we are going to introduce an implementation of the selector.

%indicating the best solution in different situations can be proposed as shown in Fig. \ref{fig.framework}, which can guide the
%Therefore, whether dense correspondences are helpful highly depends on the environment.

\subsection{Learning-based solution selection}
\begin{figure}[tbp]
    \centering
    \includegraphics[width=0.5\textwidth]{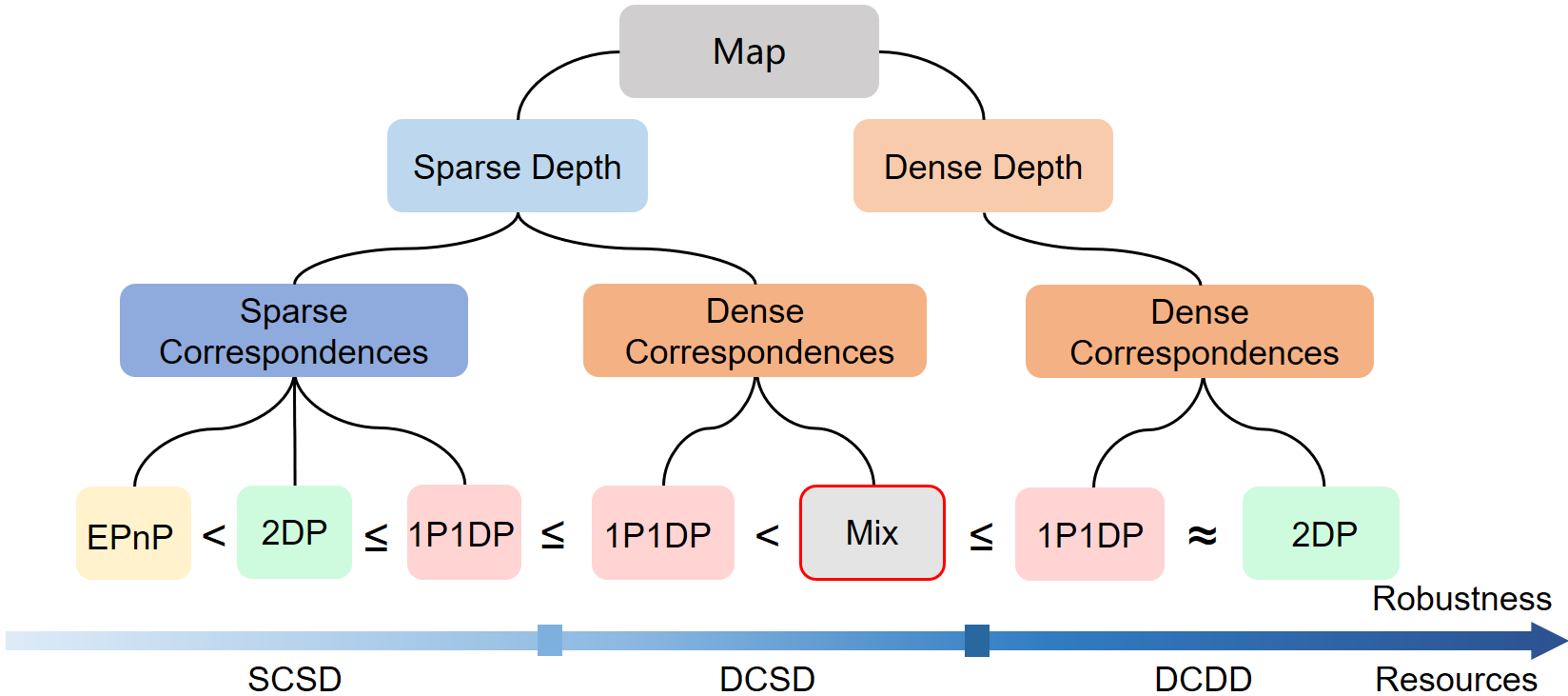}
    \caption{Robust localization pipeline under different settings. The robustness increases from left to right as analysed in theory. The computation and storage resources also increase due to the density of the correspondences and depth.}
    \label{fig.framework}
  \vspace{-0.3cm}
\end{figure}

%Above analysis indicates that the
Following the result above, we implement a learning based solution selector based on the reliability of depth in map. The solution is denoted as Mix in Fig. \ref{fig.framework}. The network takes a three-channel image as the input including depth, depth uncertainty and observation number for the features in the reference image, and predict the utilization of 1P1DP or 2DP given that reference image, as shown in Fig. \ref{fig.example}. ResNet-101 \cite{he2016deep} is selected as the backbone followed by a fully connected layer with two outputs. For data construction, we run 1P1DP and 2DP for each training image, and use the solution giving better accuracy as the label. Some examples of the training data are shown in Fig. \ref{fig.example}.
%\subsection{Situation with dense map}
%When dense map is available, the difference of the situation with sparse match and sparse map is that the number of feature correspondences is significantly increased. Then the impact of low reliable depth features in 2dp computation will be reduced and even ignored. Therefore, 2dp can be the best choice with better accuracy and robustness.

\section{Experimental Results}
To validate over the state-of-the-art algorithms, we perform simulation experiments with generated synthetic data and real-world experiments with public indoor localization dataset to evaluate: i) the accuracy of the proposed 1P1DP and MC1P1DP in the simulated mono-camera and multi-camera system, ii) the robustness with increasing outlier rate and decreasing reliable depth rate, iii) the success rate of the different minimal solutions and learning-based solution selection mechanism in real-world indoor localization.

\subsection{Simulation experiments}
By fixing one reference camera coordinate system as the world, we create the synthetic world by randomly generating points in $[-8,8]^3$ cube. Then random planar motion is generated by creating the pose of query camera with random translation in range of $[-2,2]$ along $x$-axis and $z$-axis, and random rotation in range of $[-\pi,\pi]$ around $y$-axis. The focal length of the virtual camera is fixed as 800 with the resolution of $1280\times960$ and the principal point as $(640,480)$. For multi-camera system, three cameras with a viewing angle difference of 60 degree and a distance difference of 0.25m are bounded together. In each experimental test, 50 visible points are sampled for each camera and projected in both reference and query view, which is similar as in \cite{kneip2013using}. For evaluation, given the ground truth of the pose of query camera $[R_{gt}|t_{gt}]$, the rotation error of the estimated pose $[R|t]$ is computed as $\arccos (0.5Tr({R}R_{gt}^T)-0.5)$ in degree, and the translation error is $|{t}-t_{gt}|$ in meter \cite{sattler2018benchmarking}.

\textbf{Accuracy}: To evaluate the accuracy of different algorithms, we generate Gaussian noise with zero mean and increasing standard deviation from 0 to 5 pixels to the 2D features. The Gaussian noise on the depth of the 3D points is also added and the standard deviation is fixed as 0.05m. For test on mono-camera system, besides the 2DP \cite{jiao20202}, the generally utilized 6DoF algorithms in real pose estimation tasks EPnP \cite{lepetit2009epnp} as well as the latest improved version AP3P \cite{ke2017efficient} are considered. For multi-camera pose estimation algorithms, GP3P \cite{kneip2013using} and MC2DP \cite{jiao20202} are compared.

\begin{figure*}[tbp]
  \centering
  \subfigure[translation error]{
            \includegraphics[width=0.22\textwidth]{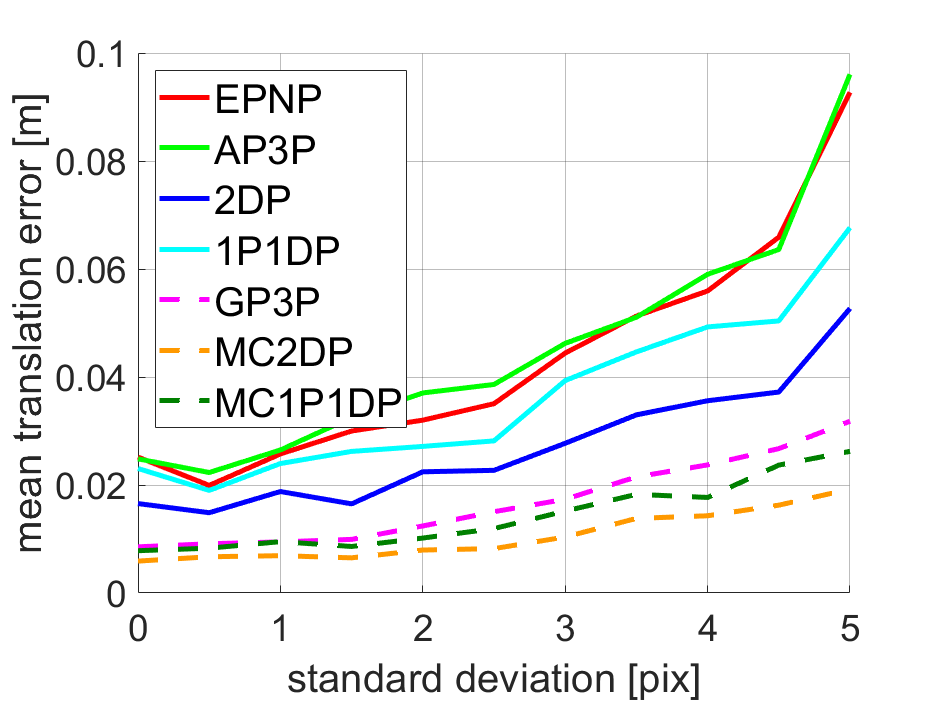}
        }
  \subfigure[AP3P]{
            \includegraphics[width=0.22\textwidth]{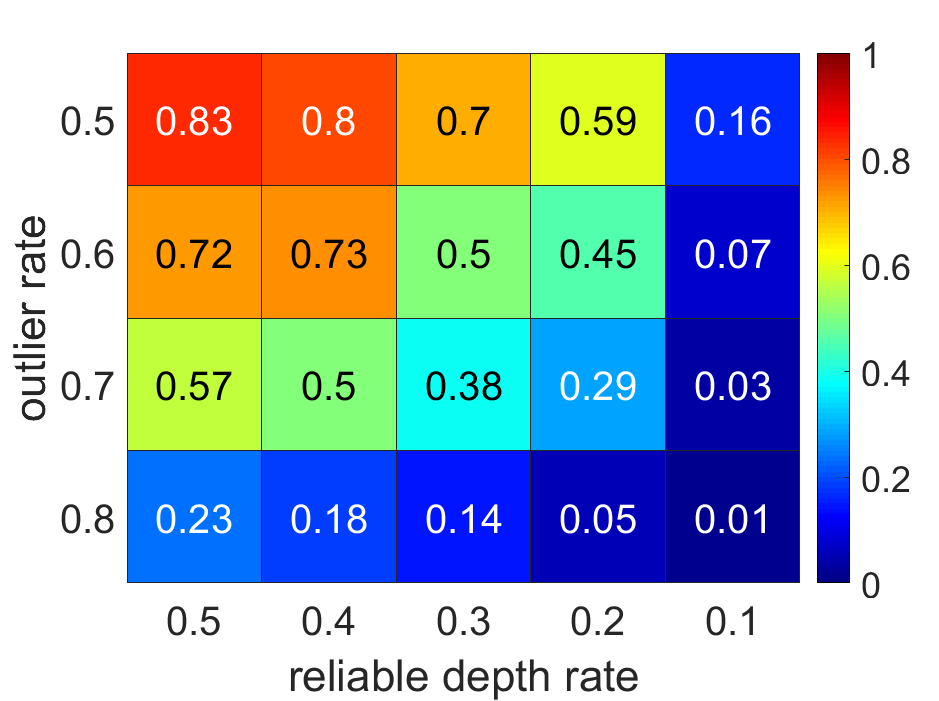}
        }
  \subfigure[2DP]{
            \includegraphics[width=0.22\textwidth]{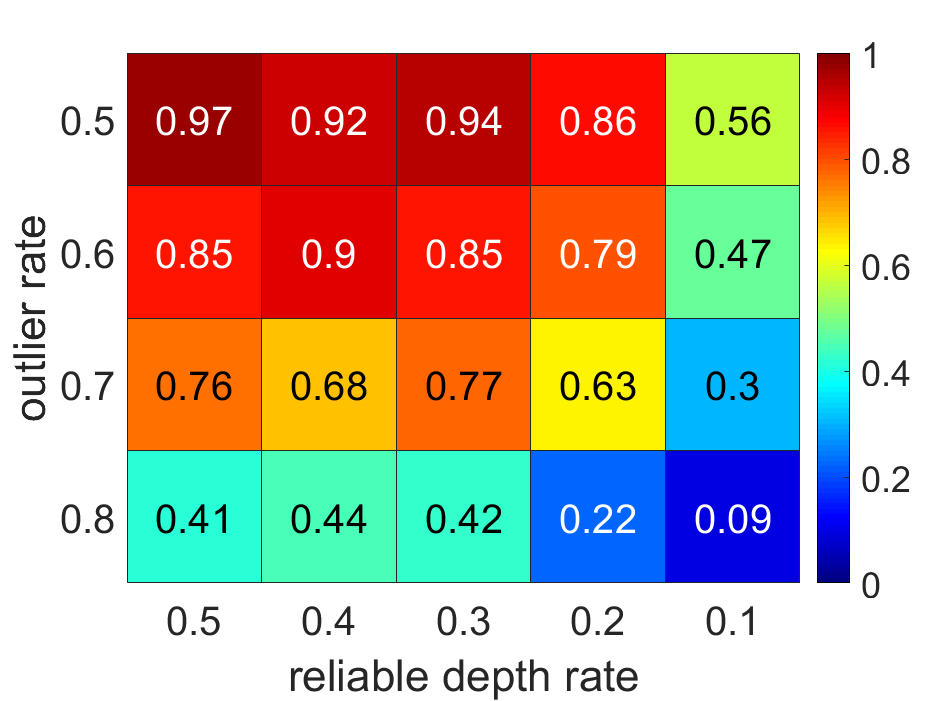}
        }
  \subfigure[1P1DP]{
            \includegraphics[width=0.22\textwidth]{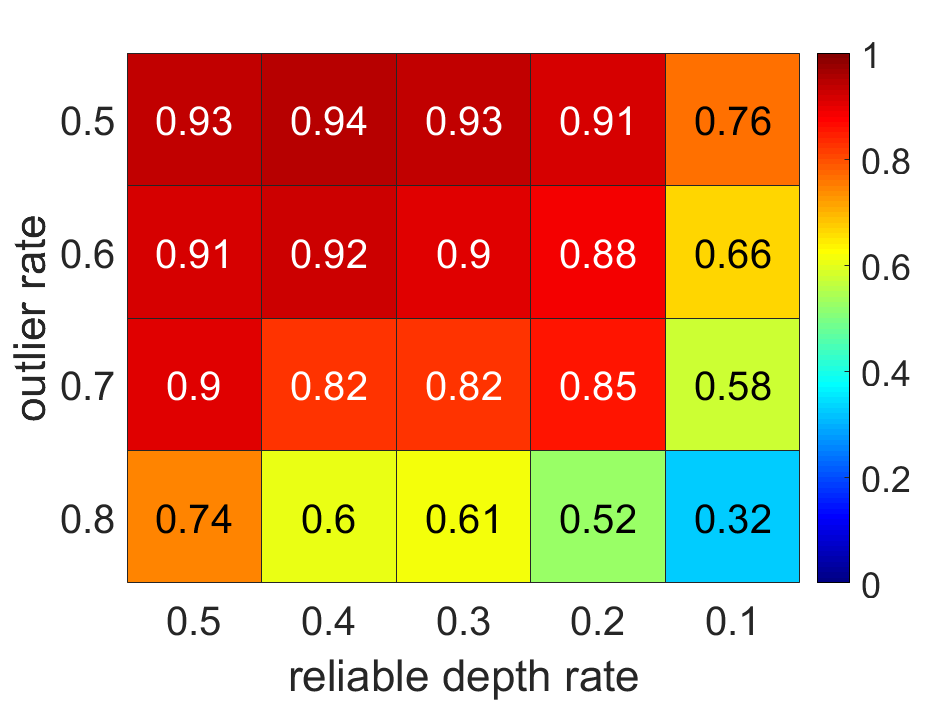}
        }
  \subfigure[rotation error]{
            \includegraphics[width=0.22\textwidth]{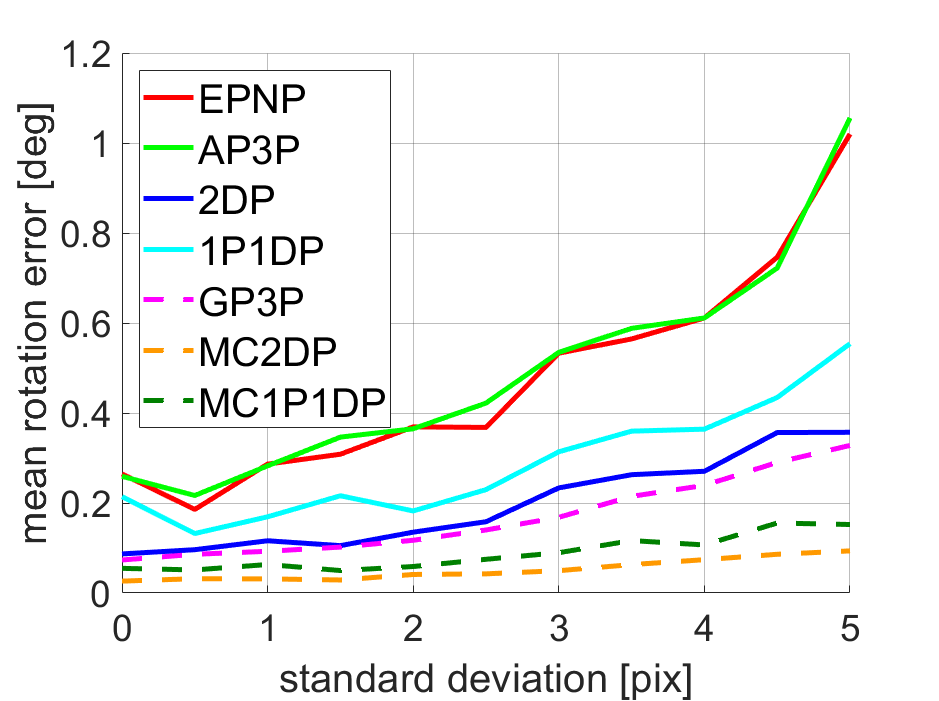}
        }
  \subfigure[GP3P]{
            \includegraphics[width=0.22\textwidth]{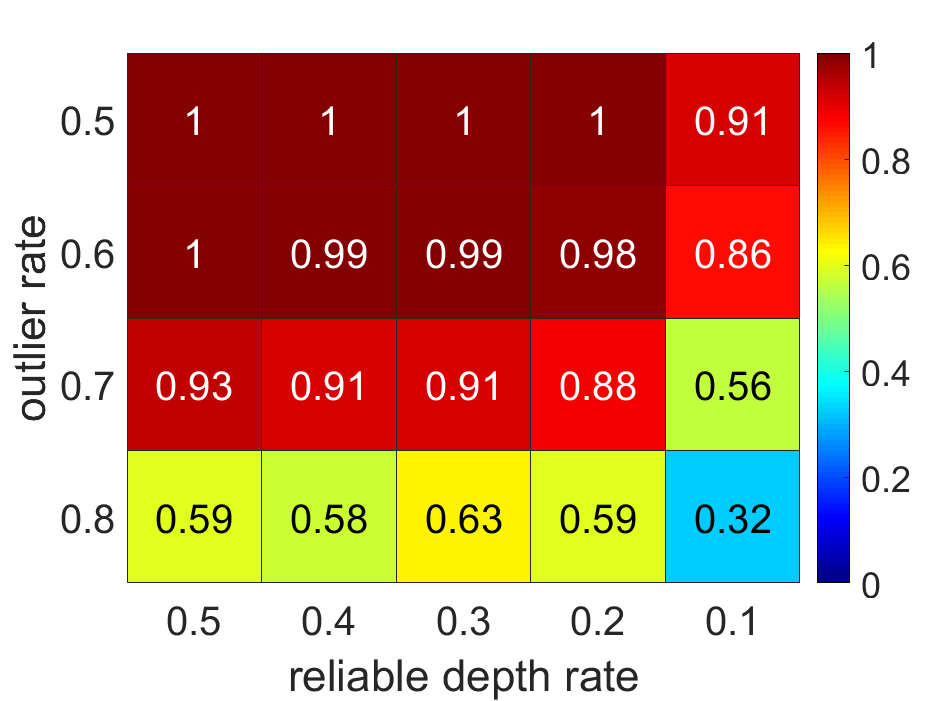}
        }
  \subfigure[MC2DP]{
            \includegraphics[width=0.22\textwidth]{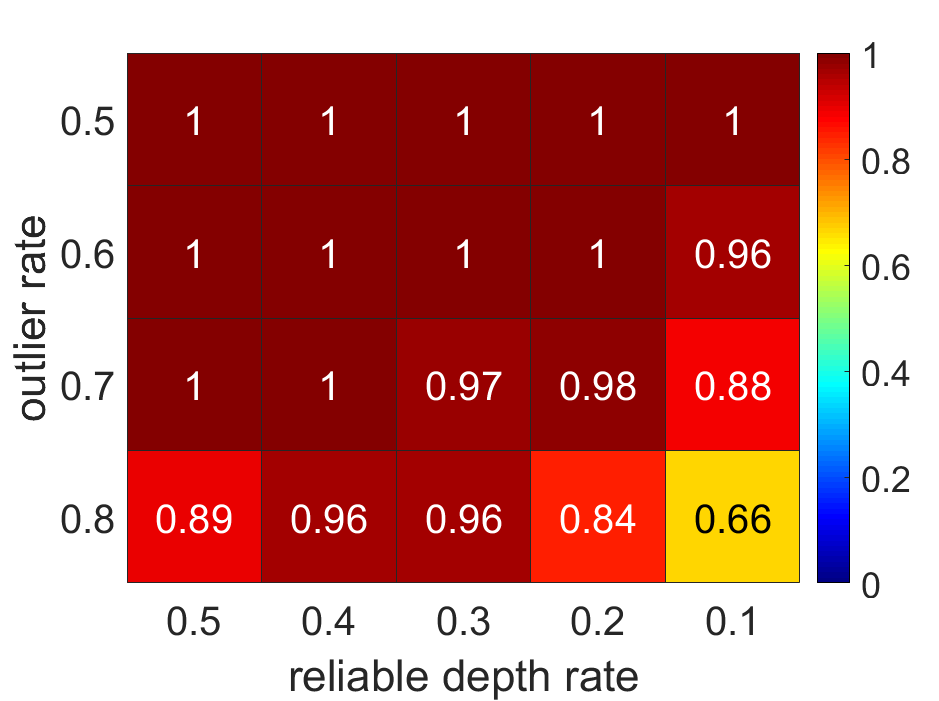}
        }
  \subfigure[MC1P1DP]{
            \includegraphics[width=0.22\textwidth]{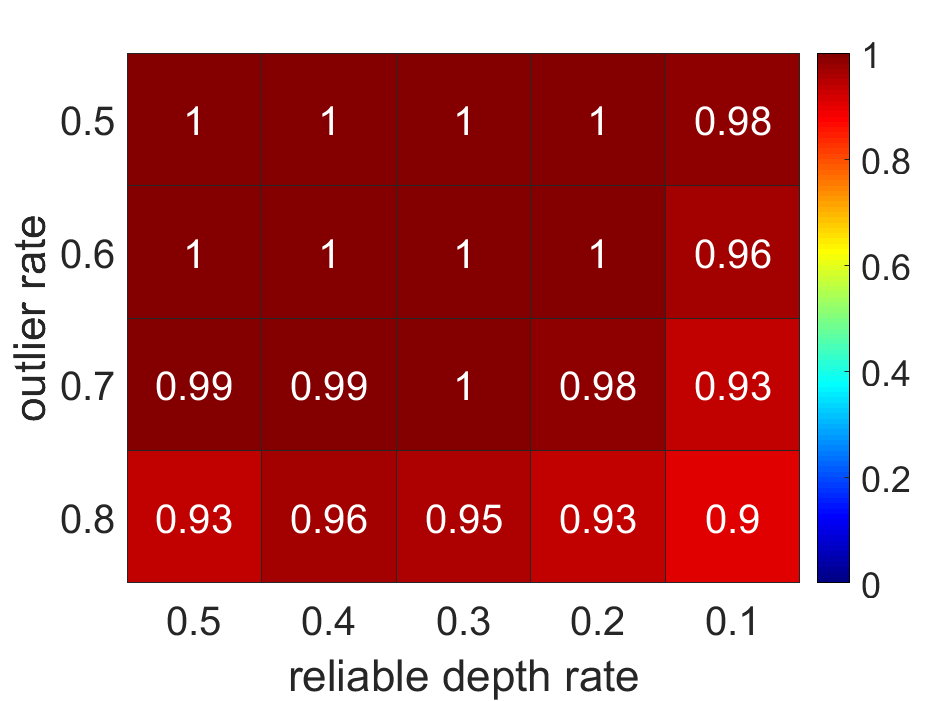}
        }
  \caption{(a) and (e): The accuracy comparison of mono-camera and multi-camera algorithms with increasing 2d noise. (b)-(d) and (f)-(h): The success rate comparison with increasing outlier rate and decreasing reliable depth rate of mono-camera and multi-camera algorithms.}
  \label{fig.accuracy}
  \vspace{-0.5cm}
\end{figure*}

We generate 100 experimental tests for each noise level. In each test, all mono-camera and multi-camera algorithms are embedded into RANSAC and $5000$ iterations are performed, then the result with the most inliers is selected as the estimated pose. The mean error of translation and rotation over all 100 tests are drawn for mono-camera and multi-camera algorithms comparison in Fig. \ref{fig.accuracy} (a) and (e). Results show that the solutions designed for reduced DoF problem achieve better accuracy, both in mono-camera and multi-camera system, as the number of correspondences required for pose estimation is reduced. And the 2DP solution which utilizes more depth information performs slightly better than proposed 1P1DP solution, which is reasonable as the constraints are stronger.

\textbf{Robustness}: Two experimental settings are designed to test the robustness of mono-camera and multi-camera algorithms: i) increasing outlier rate between all feature correspondences, ii) decreasing reliable depth rate to vary the ratio of correspondences with and without depth (the correspondence with unreliable depth is assumed to be no depth). The total number of feature correspondences for each camera is fixed as 50 and the outliers are generated by randomly associating a certain number of points with random incorrect camera poses. The outlier rate varies from 0.5 to 0.8. Meanwhile, the reliable depth rate is decreasing from 0.5 to 0.1, which means the number of correspondences with depth decreases. The default Gaussian noise with 2 pixel standard deviation on 2D measurements and 0.05m on 3D measurements are added for all cases. The localization result is assumed to be successful if the translation error is lower than 0.1m and rotation error is lower than 1 degree. Total 100 experimental tests for each level are generated and 500 iterations are performed for each algorithm in each test, and the success rate of different algorithms is counted and averaged for robustness comparison.

The results of mono-camera algorithms are shown in Fig. \ref{fig.accuracy} (b)-(d) and multi-camera algorithms in (f)-(h). Results show that the proposed 1P1DP performs better than 2DP and other depth correspondence based solutions, as more outliers and less depth measurements present. In addition, the success rate of multi-camera algorithms is higher than that of mono-camera algorithms, as more correspondences are available which provides more positive candidates.

\subsection{Real-world experiments}
OpenLORIS dataset \cite{shi2020we} is employed to evaluate the feasibility of the proposed algorithm in real world planar moving robot visual localization. The dataset is collected by a wheeled robot equipped with a RealSense D435i and a RealSense T265. The dataset contains challenging lifelong variations including viewpoints, illuminations, dynamic objects changes and human occlusions. The RGBD data from RealSense D435i is utilized for evaluation in this paper. We test the success rate of different algorithms in all situations and show the result of the proposed solution selector, as the validation of the proposed pipeline.

\vspace{-1mm}
\begin{table*}[tbp]
\caption{Success rate comparison of multi-camera algorithms.}
\vspace{-0.3cm}
\begin{center}
\resizebox{1.0\textwidth}{!}{
\begin{threeparttable}
\begin{tabular}{llcccccc}
\Xhline{1pt}
\multirow{3}{*}{Task} & session      & home2                     & home4                     & home5                   & corridor2               & corridor3             & corridor4                 \\
&m                            & 0.25 / 0.5 / 1.0      & 0.25 / 0.5 / 1.0      & 0.25 / 0.5 / 1.0      & 0.25 / 0.5 / 1.0      & 0.25 / 0.5 / 1.0      & 0.25 / 0.5 / 1.0      \\
&degree                          & 5 / 5 / 5             & 5 / 5 / 5             & 5 / 5 / 5             & 5 / 5 / 5             & 5 / 5  /5             & 5 / 5 / 5             \\
\Xhline{1pt}
\multirow{3}{*}{SCSD}  &GP3P         & 68.81 / 80.69 / 82.79     & 51.06 / 65.81 / 70.36     & 77.02 / 79.20 / 79.20   & 11.42 / 33.91 / 57.66   & 2.29 / 6.67 / 14.39   & 18.16 / 30.92 / 35.63     \\
&MC2DP        & 77.25 / 84.39 / \textcolor{black}{85.69}     & 71.30 / 83.33 / 87.41     & \textcolor{black}{98.33} / \textcolor{black}{98.46} / \textcolor{black}{98.46}   & 18.15 / \textcolor{black}{43.14} /   \textcolor{black}{62.64} & \textcolor{black}{3.53} /   9.34 / 19.53 & \textcolor{black}{22.18} / \textcolor{black}{36.43} /   43.10   \\
&MC1P1DP (ours)      & \textcolor{black}{77.52} / \textcolor{black}{84.56} / \textcolor{black}{85.69}     & \textcolor{black}{72.85} / \textcolor{black}{84.64} / \textcolor{black}{87.74}     & \textbf{99.87} / \textbf{99.87} / \textbf{99.87}   & \textcolor{black}{18.84} /   42.97 / 62.15 & \textcolor{black}{3.62} / \textcolor{black}{9.62} / \textcolor{black}{21.63}   & \textcolor{black}{20.28} /   36.37 / \textcolor{black}{43.52}   \\
\multirow{2}{*}{DCSD} &DCSD-MC1P1DP (ours) & \textcolor{black}{77.65} / \textcolor{black}{84.82} / \textcolor{black}{85.89}     & \textcolor{black}{73.07} / \textcolor{black}{86.31} / \textcolor{black}{89.53}     & \textbf{99.87} / \textbf{99.87} / \textbf{99.87}   & \textcolor{black}{18.61} / \textcolor{black}{43.54}   / \textcolor{black}{62.96} & 3.48 / \textcolor{black}{9.62} / \textcolor{black}{21.15}   & 20.17 / \textcolor{black}{36.74}   / \textcolor{black}{43.67}   \\
&MCMix (ours) & \textbf{78.89} / \textbf{85.22} / \textbf{86.69}     & \textbf{74.78} / \textbf{86.75} / \textbf{90.23}     & \textbf{99.87} / \textbf{99.87} / \textbf{99.87}   & \textbf{18.98} / \textbf{43.60}   / \textbf{63.30} & \textbf{4.48} / \textbf{11.05} / \textbf{22.49}   & \textbf{23.35} / \textbf{39.33}   / \textbf{45.53}   \\ \hline
\multirow{3}{*}{DCDD}                     & GP3P                              & 67.24 / 82.35 / 84.16     & 57.16 / 76.51 / 81.54     & 88.06 / 89.73 / 89.73   & 15.39 / 38.42 / 58.12   & 3.33 /  11.62 / 27.35  & 29.75 / 45.53 / 53.57     \\
    & MC2DP                             & 77.79 / 85.92 / 87.06     & 74.26 / 88.82 / 94.03     & 99.23 / 99.36 / 99.36   & 25.77 / 46.33 / 62.67   & \textbf{9.72} /   \textbf{20.44} / \textbf{32.78} & 34.46 / 47.49 / 54.37     \\
    & MC1P1DP (ours)                           & \textbf{77.95} / \textbf{86.22} /  \textbf{88.06}    & \textbf{74.50} / \textbf{89.24} / \textbf{94.22}     & \textbf{99.87} / \textbf{100} / \textbf{100}       & \textbf{27.78} / \textbf{48.09} / \textbf{62.90}   & 8.86 / 18.20 / 30.35   & \textbf{36.05} / \textbf{48.28} / \textbf{54.53}     \\ \Xhline{1pt}
    & & & & & & & \\
\Xhline{1pt}
\multirow{3}{*}{Task} &  session      & corridor5                 & office3                   & office4                 & office5                 & office7               & cafe1                     \\
&m                            & 0.25 / 0.5 / 1.0      & 0.25 / 0.5 / 1.0      & 0.25 / 0.5 / 1.0      & 0.25 / 0.5 / 1.0      & 0.25 / 0.5 / 1.0      & 0.25 / 0.5 / 1.0      \\
&degree                          & 5 / 5 / 5             & 5 / 5 / 5             & 5 / 5 / 5             & 5 / 5 / 5             & 5 / 5  /5             & 5 / 5 / 5             \\
\Xhline{1pt}
\multirow{3}{*}{SCSD}  &GP3P         & 37.00 / 59.57 / 77.78     & 88.33 / 88.33 / 88.33     & 78.97 / 79.77 / 79.77   & 95.17 / 95.29 / 95.29   & 98.86 / 98.95 / 98.95 & 36.87 / 71.84 / 86.30     \\
&MC2DP        & 37.99 / 62.84 / 80.29     & \textcolor{black}{90.28} /   90.28 / \textcolor{black}{90.56}   & \textcolor{black}{82.99} /   85.09 / 85.21 & \textcolor{black}{98.05} / \textcolor{black}{98.43} / \textcolor{black}{98.43}   & \textcolor{black}{99.47} / \textcolor{black}{99.47} / \textcolor{black}{99.47} & \textcolor{black}{42.21} / \textcolor{black}{74.88}   /   86.71 \\
&MC1P1DP (ours)      & \textcolor{black}{40.70} / \textcolor{black}{63.29}   /   \textcolor{black}{80.36} & \textcolor{black}{90.83} / \textbf{91.11}   /   \textcolor{black}{91.11} & \textcolor{black}{82.64} / \textcolor{black}{85.40} / \textcolor{black}{85.52}   & 96.60 / 96.79 / 96.79   & \textcolor{black}{99.56} / \textcolor{black}{99.56} / \textcolor{black}{99.56} & \textcolor{black}{38.17} /   73.65 / \textbf{86.89}   \\
\multirow{2}{*}{DCSD} &DCSD-MC1P1DP (ours) & \textcolor{black}{40.68} / \textcolor{black}{63.16} /   \textcolor{black}{80.63}   & 90.00 / \textcolor{black}{90.56} /   \textcolor{black}{90.56}   & \textcolor{black}{82.64} / \textcolor{black}{85.63}   / \textcolor{black}{85.63} & \textcolor{black}{96.66} / \textcolor{black}{96.97} / \textcolor{black}{96.97}   & 99.39 / 99.39 / 99.39 & 37.76 / \textcolor{black}{73.83} /   \textcolor{black}{86.49} \\
&MCMix (ours) & \textbf{41.18} / \textbf{64.64} / \textbf{81.50}     & \textbf{91.11} / \textbf{91.11} / \textbf{91.39}     & \textbf{83.90} / \textbf{86.78} / \textbf{86.90}   & \textbf{98.24} / \textbf{98.55}   / \textbf{98.55} & \textbf{99.63} / \textbf{99.63} / \textbf{99.63}   & \textbf{43.33} / \textbf{77.05}   / \textbf{86.89}   \\ \hline
\multirow{3}{*}{DCDD}                     & GP3P                              & 40.98 / 63.32 / 81.27     & 88.61 /   89.44 / 89.44   & 88.62 /   89.31 / 89.31 & 95.41 / 95.97 / 95.97   & 93.51 / 95.53 / 95.53   & 40.57 / 73.48 / 88.66     \\
      & MC2DP                             & 47.46 / 68.98 / \textbf{85.15}     & \textbf{90.00} / \textbf{91.11} / \textbf{91.11}     & \textbf{92.18} / \textbf{93.45} / \textbf{93.56}   & \textbf{97.29} / \textbf{97.48} / \textbf{97.48}   & 94.83 / \textbf{95.79}   / \textbf{95.79} & \textbf{52.99} / \textbf{78.63} / \textbf{89.23}     \\
      & MC1P1DP (ours)                           & \textbf{49.36} / \textbf{69.51} / 85.06     & 88.27 / 90.00 / 90.00     & 87.47 / 90.80 / 90.80   & 96.22 / 96.79 / 96.79   & \textbf{94.92} / \textbf{95.79} / \textbf{95.79}   & 51.23 / 78.57 / 88.70     \\ \Xhline{1pt}
\end{tabular}
\begin{tablenotes}
	\footnotesize
	\item[1] Considering the map coverage of the environment, the map sequences for each session are as follows: home1 and home3, corridor1, office1 and office2, cafe2.
    \item[2] For testing on office6, all methods achieve 100 success rate over three thresholds.
\end{tablenotes}
\end{threeparttable}}
\end{center}
\label{table.multi}
\vspace{-0.5cm}
\end{table*}

\textbf{Experimental Settings and Implementation Details}: For each query image, top 3 reference images in the map session are retrieved using NetVLAD \cite{arandjelovic2016netvlad}. Then R2D2 \cite{revaud2019r2d2} is utilized to get sparse correspondences between query and reference images and NCNet\cite{rocco2020ncnet} is utilized for dense correspondences. For map building, we triangulate scene points with the observation from multiple frames and refine the reconstruction with bundle adjustment. In evaluation of multi-camera algorithms, 5 temporal images including the query image in query session are retrieved with rotation difference larger than 20 degree and translation difference larger than 0.25 m. The aligned depth image from RGBD sensor is utilized as dense depth map in DCDD test. 500 iterations are performed for each solution in RANSAC and the estimated pose is refined with nonlinear optimization for better accuracy.

\begin{figure}[tbp]
    \centering
    \includegraphics[width=0.5\textwidth]{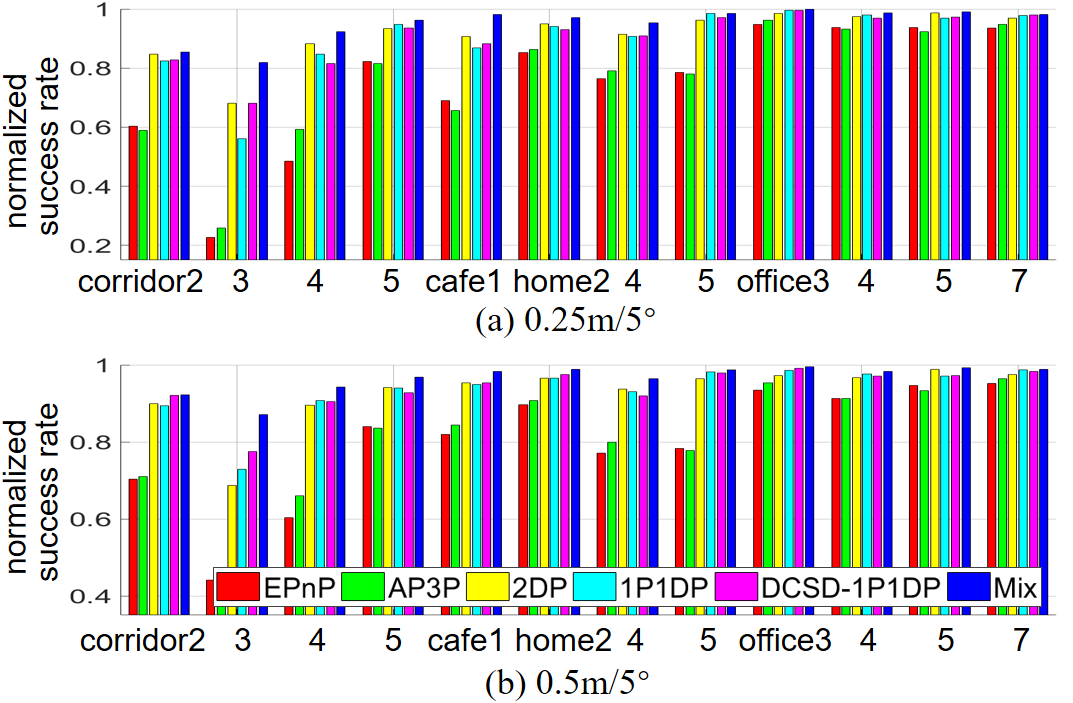}
    \caption{The normalized success rate comparison of mono-camera algorithms in all sessions.}
    \label{fig.monosuc}
  \vspace{-0.5cm}
\end{figure}

For localization performance evaluation, different error thresholds are utilized to count the success rate as shown in Fig. \ref{fig.monosuc} for mono-camera and Table. \ref{table.multi} for multi-camera algorithms. For better illustration, the success rate shown in Fig. \ref{fig.monosuc} is normalized by the upper bound performance in each session. The upper bound is counted by manually selecting the best result with smallest error among all algorithms in each query case. For multi-camera algorithms, the success rate is the absolute value. In 1P1DP sampling, only map points with depth smaller than 5m is considered to be reliable depth points (DP). All other map points with larger depth as well as those without depth are considered to be points without depth (P). The threshold is chosen according to the inherent attributes of the RGBD sensor.
%For mono-camera algorithms comparison, we also count the result by manually selecting the pose with smaller error between all algorithms in each query case to represent the upper bound performance.

\begin{figure}[tbp]
    \centering
    \includegraphics[width=0.5\textwidth]{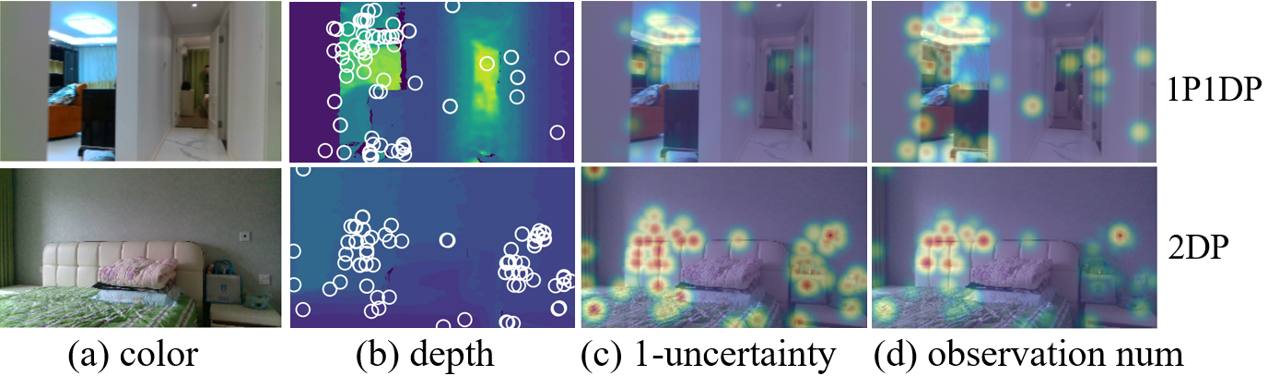}
    \caption{Examples of training data. Only sparse depth of the detected features are combined with the uncertainty and observation number to fed into the network.}
    \label{fig.example}
  \vspace{-0.3cm}
\end{figure}

Some implementation details of solution selection network are as follows. Each channel of input data is generated as a heatmap processed with Gaussian blur and the resolution is the same as RGB image as shown in Fig. \ref{fig.example}. Training set is composed of the reference images with difference in accuracy between 1P1DP and 2DP exceeding 0.25 m in translation. We train the network by SGD \cite{lecun1989backpropagation} solver with 12 batch size for 200 epochs. Then the trained selector is performed in each query to automatically select the best solution for pose estimation. The obtained success rate is denoted as Mix in mono-camera and MCMix in multi-camera system.
%Then the trained selector is performed in each query including that correspond to repeatedly retrieved and unseen reference images to automatically select the best solution for pose estimation.
% All other repeatedly observed and unseen reference images are used for testing. The result of model selection is denoted as Mix in mono-camera and MCMix in multi-camera system.
%And the standard deviation of the Gaussian function is 20 pixels.
% The ground truth of reference images are labelled as 1 if 1P1DP has higher accuracy and 0 on the contrary.
% As multiple query images may match the same reference image, only the first is utilized for training data labeling. 

\begin{figure}[tbp]
    \centering
    \includegraphics[width=0.5\textwidth]{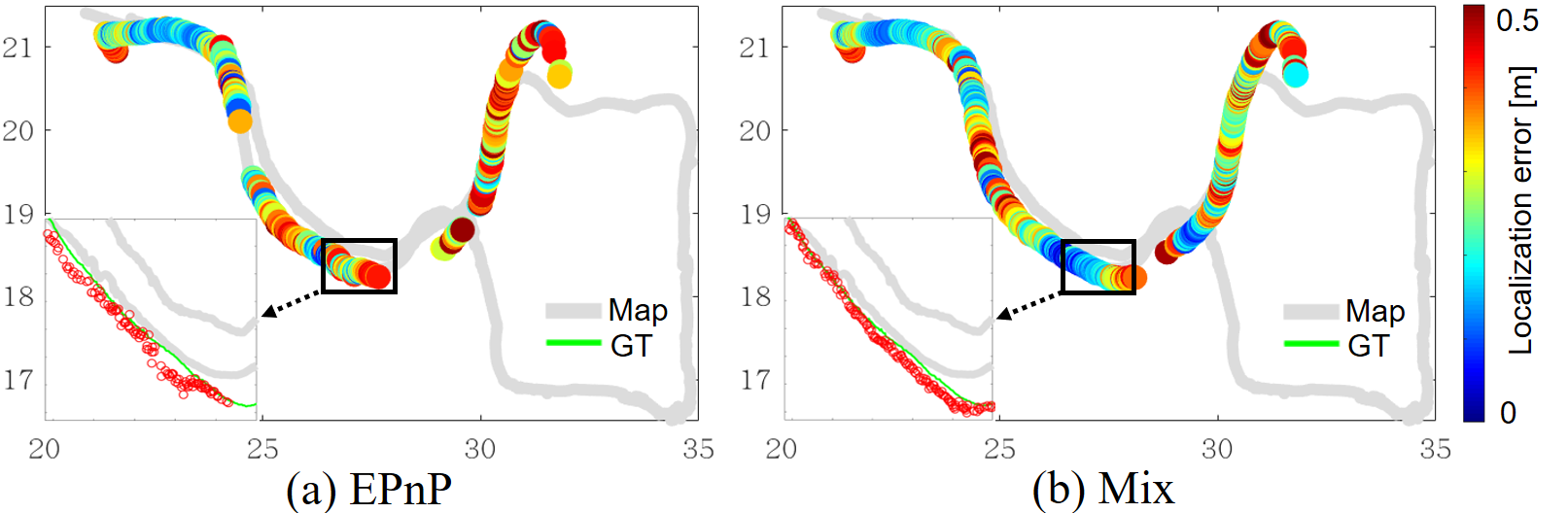}
    \caption{Localization performance comparison on \textit{cafe1}. The proposed Mix solution achieves more success localization result with higher accuracy.}
    \label{fig.traj}
  \vspace{-0.3cm}
\end{figure}

\textbf{Performance on Sparse Depth}: Results in mono-camera and multi-camera system of SCSD show that, the performance of 1P1DP and 2DP outperform the common 6DoF algorithms under all thresholds which confirms the importance of utilizing motion property into the pose estimation. Comparing result of 1P1DP and 2DP in \textit{cafe1} and \textit{corridor5}, we can find that 1P1DP shows advantage in \textit{corridor5} where the reliable depth rate is low due to the large French windows. Whereas 2DP performs better in \textit{cafe1} where reliable depth rate is relatively higher as the scene is full of texture and close to the camera. On the whole, 1P1DP achieves slightly better performance than 2DP which is more obvious in DCSD-1P1DP. The result reflect the fact that 2DP and 1P1DP can complement each other according to different environmental characteristics, which also explains the reason that the performance of the proposed solution selector is the best in all sessions. A comparison of localization performance on whole trajectory of \textit{cafe1} is also drawn in Fig. \ref{fig.traj} for clear illustration. Therefore, utilizing an appropriate selection strategy to combine the advantages of 2DP and 1P1DP can achieve the best performance in sparse depth condition, which is consistent with the analysis in Section \ref{model}.

%The success rate of 2DP under the smallest threshold are mostly better than 1P1DP, which reflects better accuracy. Whereas under larger thresholds, the 1P1DP shows advantages which is more obvious in DCSD, indicating better robustness. And the performance of the proposed model selection is the best under all thresholds in all sessions, reflecting the fact that 2DP and 1P1DP can complement each other in different environments. Note that the proposed selection algorithm shows more advantage in challenging environments, such as \textit{corridor} in which the illumination variations is significant and \textit{cafe} where human interference is frequent. Therefore, utilizing an appropriate selection strategy to combine the advantages of 2DP and 1P1DP can achieve the best performance in sparse depth condition, which is consistent with the analysis in Section \ref{model}.

\textbf{Discussion about Dense Depth}: In the environment with larger changes (significant illumination variations in \textit{corridor}, frequent human activities in \textit{home} and \textit{cafe}), the performance of all methods with dense depth are obviously better than that with sparse depth. However, in the relatively stable \textit{office} environment, the performance with dense depth slightly drops. That is reasonable in two aspects: i) the dense depth from the RGBD sensor is not accurate as the optimized depth in sparse map ii) sparse features perform well in this environment, then dense features will not provide more significant correspondences. And MC2DP performs better in most cases which indicates the MC2DP can be the best choice with better accuracy and robustness when dense depth map is reliable.


\section{Conclusions}

In this paper, a novel minimal solution with both minimal number of feature correspondences and minimal depth information namely 1P1DP is proposed. Embedded with the solution, all the matched feature correspondences between query and reference image (with or without depth) can be integrated for pose estimation, which maximizes the sample set in RANSAC framework. Furthermore, the solution enables the combination of dense correspondences and sparse depth, taking advantages of dense features about changing environment while maintaining limited computational requirement for sparse map building. In the future, the strategy for selecting reliable depth features for 1P1DP will be optimized.


%%%%%%%%%%%%%%%%%%%%%%%%%%%%%%%%%%%%%%%%%%%%%%%%%%%%%%%%%%%%%%%%%%%%%%%%%%%%%%%%

%%%%%%%%%%%%%%%%%%%%%%%%%%%%%%%%%%%%%%%%%%%%%%%%%%%%%%%%%%%%%%%%%%%%%%%%%%%%%%%%

%%%%%%%%%%%%%%%%%%%%%%%%%%%%%%%%%%%%%%%%%%%%%%%%%%%%%%%%%%%%%%%%%%%%%%%%%%%%%%%%

%\section*{ACKNOWLEDGMENT}

%%%%%%%%%%%%%%%%%%%%%%%%%%%%%%%%%%%%%%%%%%%%%%%%%%%%%%%%%%%%%%%%%%%%%%%%%%%%%%%%

%References are important to the reader; therefore, each citation must be complete and correct. If at all possible, references should be commonly available publications.

\bibliographystyle{ieeetr} %% setting the cite style
\bibliography{dense}

\end{document}